\documentclass[10pt,twocolumn,letterpaper]{article}

\def\authorBlock{
    Hwan Heo \qquad
    Jangyeong Kim \qquad
    Seongyeong Lee \\
    Jeong A Wi  \qquad
    Junyoung Choi  \qquad
    Sangjun Ahn \thanks{corresponding author.} \\ 
    \centering
    Graphics AI Lab, NC Research \\
    {\tt\small \{hwanheo, jangyeongk, seongyeong2, jaywi, jychoi13, sjahn21\}@ncsoft.com}
}

\newif\ifreview 
\newif\ifarxiv \newcommand{\arxiv}{\arxivtrue}
\newif\ifcamera 
\newif\ifrebuttal 

\arxiv

\ifreview \usepackage[review]{cvpr} \fi
\ifarxiv \usepackage[pagenumbers]{cvpr} \fi
\ifrebuttal \usepackage[rebuttal]{cvpr} \fi
\ifcamera \usepackage{cvpr} \fi

\usepackage{ifthen}
\usepackage{graphicx}	
\usepackage{amsmath}	
\usepackage{amssymb}	
\usepackage{booktabs}
\usepackage{times}
\usepackage{microtype}
\usepackage{epsfig}
\usepackage[table,xcdraw,dvipsnames]{xcolor}
\usepackage{caption}
\usepackage{float}
\usepackage{placeins}
\usepackage{color, colortbl}
\usepackage{stfloats}
\usepackage{enumitem}
\usepackage{tabularx}
\usepackage{xstring}
\usepackage{multirow}
\usepackage{xspace}
\usepackage{url}
\usepackage{subcaption}
\usepackage{xcolor}
\usepackage[hang,flushmargin]{footmisc}

\ifcamera \usepackage[accsupp]{axessibility} \fi

% ...

\ifarxiv  \fi

\newcommand{\R}[1]{{%
    \textbf{%
        \ifstrequal{#1}{1}{\textcolor{red}{R#1}}{%
        \ifstrequal{#1}{2}{\textcolor{blue}{R#1}}{%
        \ifstrequal{#1}{3}{\textcolor{magenta}{R#1}}{%
        \ifstrequal{#1}{4}{\textcolor{teal}{R#1}}{%
                           \textcolor{cyan}{R#1}%
        }}}}%
    }%
}}

\usepackage{xr-hyper}

\makeatletter
\newcommand*{\addFileDependency}[1]{
  \typeout{(#1)}
  \@addtofilelist{#1}
  \IfFileExists{#1}{}{\typeout{No file #1.}}
}

\makeatother

\definecolor{cvprblue}{rgb}{0.21,0.49,0.74}
\usepackage[pagebackref,breaklinks,colorlinks,citecolor=cvprblue]{hyperref}
\usepackage[capitalize]{cleveref}
\crefname{section}{Sec.}{Secs.}
\crefname{table}{Table}{Tables}
\crefname{figure}{Fig.}{Figs.}

\ifarxiv \crefname{appendix}{App.}{Apps.}
\else \crefname{appendix}{Suppl.}{Suppls.} \fi

\begin{document}
\title{
CaPa: Carve-n-Paint Synthesis for Efficient 4K Textured Mesh Generation
}
\makeatletter
\let\@oldmaketitle\@maketitle%
\renewcommand{\@maketitle}{\@oldmaketitle%
\centering
 \vskip -5em 
 \authorBlock 
 \vspace{0.5em} 
    \includegraphics[width=\textwidth]{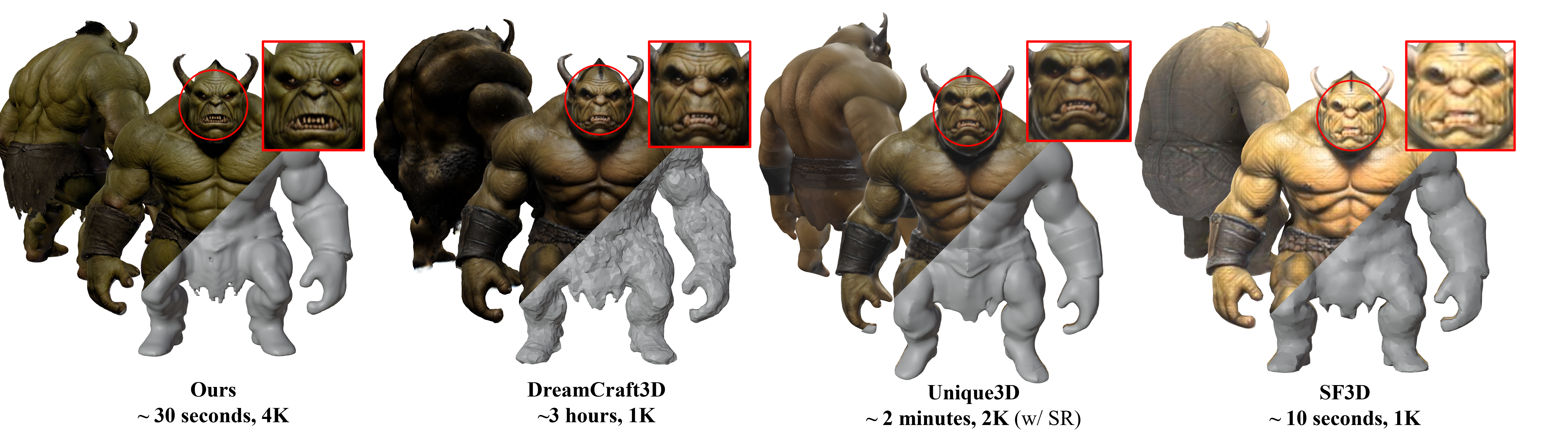}
     \captionof{figure}{
     Comparison of \textit{mesh} quality with state-of-the-art image-to-3D methods.
     \textbf{CaPa} can generate a hyper-quality \textit{textured mesh} in under 30 seconds, providing 3D assets ready for commercial applications such as games, movies, and VR/AR.
    }
    \label{fig:teaser}
    \bigskip}
\makeatother
\maketitle
\begin{abstract}
\vspace{-0.5em}
The synthesis of high-quality 3D assets from textual or visual inputs has become a central 
objective in modern generative modeling. 
Despite the proliferation of 3D generation algorithms, they frequently grapple with challenges such as multi-view inconsistency, slow generation times, low fidelity, and surface reconstruction problems. 
While some studies have addressed some of these issues, a comprehensive solution remains elusive.
In this paper, we introduce \textbf{CaPa}, a carve-and-paint framework that generates high-fidelity 3D assets efficiently. 
CaPa employs a two-stage process, decoupling geometry generation from texture synthesis. 
Initially, a 3D latent diffusion model generates geometry guided by multi-view inputs, ensuring structural consistency across perspectives. 
Subsequently, leveraging a novel, model-agnostic Spatially Decoupled Attention, the framework synthesizes high-resolution textures (up to 4K) for a given geometry. 
Furthermore, we propose a 3D-aware occlusion inpainting algorithm that fills untextured regions, resulting in cohesive results across the entire model. 
This pipeline generates high-quality 3D assets in less than 30 seconds, providing ready-to-use outputs for commercial applications.
Experimental results demonstrate that CaPa excels in both texture fidelity and geometric stability, establishing a new standard for practical, scalable 3D asset generation.
\end{abstract}
\section{Introduction}
\label{sec:intro}
The demand for scalable, high-quality 3D assets is rapidly growing across industries such as gaming, film, and VR/AR.
While recent advancements in machine learning have led to remarkable achievements in text and image generation, extending these successes to 3D content creation has been challenging.
The high-dimensional nature of 3D data, coupled with a scarcity of diverse and high-quality datasets, has constrained the progress of 3D generative models and created unique challenges in achieving efficiency and fidelity.
 
Despite these inherent obstacles, researchers have made considerable efforts to adapt generative techniques for 3D, resulting in unprecedented growth and exploration.
A prominent approach involves lifting 2D diffusion models to 3D through Score Distillation Sampling (SDS)~\cite{poole2023dreamfusion, 23sjc}.
However, SDS often encounters limitations such as the Janus problem (texture discrepancies across views), difficulties in achieving high-quality mesh, and prolonged generation times.
These shortcomings arise largely because of the overreliance on indirect 3D information from 2D models, which hampers both the efficiency and fidelity of generated outputs.
 
In response, recent advances in Large Reconstruction Models (LRM) ~\cite{hong2024lrm, zhang24gslrm} have resolved the slow speed issue by leveraging autoregressive generative models~\cite{transformer}.
While promising, these methods continue to face trade-offs, particularly in output quality.
Moreover, most 3D generation methods rely on NeRF~\cite{mildenhall2020nerf} or Gaussian Splatting~\cite{kerbl3Dgaussians} as their base representation, which has challenges for accurate surface reconstruction. 
When converting their representations into usable forms (\textit{i.e., mesh}), quality degradation occurs, remaining a key bottleneck for downstream tasks.

To address these challenges, we introduce \textbf{CaPa}, a two-stage \textit{\textbf{Ca}rve-n-\textbf{Pa}int} framework designed to generate high-quality, clean 3D meshes efficiently. 
By decoupling geometry generation and texture synthesis, CaPa enhances both flexibility and performance at each stage, allowing for precise reconstruction of the mesh and detailed texture output.

In the geometry generation stage, CaPa leverages a 3D latent diffusion model.
It focuses on producing an occupancy field that is not only structurally consistent across multiple views but also mesh-compatible. 
This process utilizes multi-view inputs to guide the generation, ensuring precise geometry that seamlessly integrates with the subsequent texture synthesis stage. The resulting high-quality mesh provides a stable foundation, crucial for achieving cohesive textures.

In the texture synthesis stage, we employ a novel \textit{Spatially Decoupled Cross Attention}, which generates high-resolution textures (up to 4K). 
By isolating and processing view-specific features within a unified diffusion framework, this method effectively resolves multi-view inconsistencies, such as the Janus problem.
Notably, its model-agnostic approach does not require architectural modifications or extensive retraining.
This design integrates smoothly with large pre-trained models like SDXL~\cite{podell2024sdxl}, 
outperforming previous approaches.
As a result, it produces geometry-aligned, cohesive, and high-fidelity textures as shown in Figure~\ref{fig:teaser}.

Additionally, to further enhance texture completeness, CaPa introduces a 3D-aware occlusion inpainting algorithm. This algorithm efficiently fills untextured regions by generating a UV map that preserves surface locality, enabling seamless inpainting while minimizing visible seams.

Comprehensive experiments validate the effectiveness of CaPa, demonstrating significant improvements in both texture fidelity and geometric stability over existing methods. 
CaPa consistently delivers high-fidelity polygonal meshes in a fraction of the time, setting a new standard for practical and scalable 3D asset generation.

The core contributions of this work are summarized as:
\begin{itemize}
    \item \textbf{Mesh-Optimized 3D Generation Pipeline}: Separating geometry and texture generation, enabling effective meshing and enhancing the practicality of 3D asset generation.
    \item \textbf{Spatially Decoupled Cross Attention}: A model-agnostic, training-free solution to generate 4K textures without additional training, super-resolution, and janus problem. 
    \item \textbf{3D-aware Fast Occlusion Inpainting}: A robust inpainting solution for handling occluded regions in 3D textures, reducing visible seams, and preserving texture fidelity.
    \item \textbf{State-of-the-art Results}: CaPa achieves significantly higher texture and geometry fidelity, marking a major step forward in practical 3D asset generation.
\end{itemize}
\section{Related Work}
\label{sec:related}

\subsection{3D Asset Generation}
\label{sec2.1}
Given the recent advancements in neural rendering~\cite{mildenhall2020nerf} and diffusion-based generative models, researchers have explored this success to the 3D generation area. 
A notable approach is Score-Distillation Sampling (SDS)~\cite{poole2023dreamfusion, 23sjc}, which utilizes 2D diffusion priors to generate 3D assets. 
Several subsequent studies~\cite{wang2023prolificdreamer, li2024sweetdreamer, liang24lucid, tang2024dreamgaussian, taoran2024gaussiandreamer, sun2024dreamcraftd} have provided a more robust theoretical foundation, post-processing, or better visual fidelity. 
Despite these advances, these methods' reliance on Neural Radiation Fields (NeRF) as the underlying 3D representation results in slow generation times and poor quality after mesh conversion.

Another prominent limitation of SDS is the ``Janus problem," where artifacts such as multiple faces appear. 
To mitigate this issue, researchers have explored enhancing diffusion models with a multi-view generation~\cite{shi2024mvdream, wang2023imagedream, liu23zero123} by adjusting U-Net to interact across these views. 
Although these methods largely resolved the Janus problem, the SDS's progressive updating still results in slower generation times. 
To accelerate the slow speed, certain studies~\cite{lu2024direct, wu2024unique3d} have proposed generating normal maps from multi-view images and employing differentiable mesh updates or sparse view neural reconstruction.
Nevertheless, this approach often leads to unstable geometry, or low-fidelity textures (refer to the second, third column of Figure~\ref{fig:main_qual}).

\subsection{3D-Native Reconstruction Models}
\label{sec2.2}
Recently, Large Reconstruction Models (LRMs)~\cite{hong2024lrm, zhang24gslrm, tang24lgm, siddiqui2024assetgen, wang2024crm, instantmesh}, a transformer models for directly generating 3D assets have emerged. 
LRMs predict the NeRF or Gaussian Splatting parameters directly through feed-forward inferences. 
These approaches significantly reduce generation times and produce 3D assets in near real-time. 
However, scaling these methods to higher resolutions remains challenging, with most state-of-the-art LRMs currently limited to low-fidelity outputs (see the fourth sample in Fiure~\ref{fig:teaser}).

Recently, the research community has explored alternative approaches for 3D \textit{shape} generation through 3D latent diffusion, diverging from the LRM paradigm. 
Michelangelo~\cite{zhao23Michelangelo} introduced VAE for 3D latent representation, establishing an aligned latent space for images, text, and 3D data. 
Several following works~\cite{Zhang24clay, wu24direct3d, li2024craftsman} enhance its geometry understanding and address its limitations in diversity. 
Concurrently, CLAY~\cite{Zhang24clay} demonstrated that 3D geometry generation, particularly using DiT~\cite{peebles23dit} (Diffusion Transformer), could scale similar to other large foundational models in deep learning. 
However, these models focus primarily on shape generation. 
To synthesize textures, they rely on external texturing methods~\cite{cao23texfusion, liu2023syncmvd, chen23text2tex} or multi-view diffusion. 
Additionally, all the aforementioned methods are time-consuming, and aligning textures with generated shapes remains difficult. 

\subsection{Texture Generation}
\label{sec2.3}
Numerous texture generation algorithms have recently been developed with the success of 2D generative diffusion models. 
Common approaches~\cite{chen23text2tex,texture23} utilize iterative processes, first rendering depth maps and then generating corresponding images for multiple viewpoints. 
However, these methods mainly encounter multi-view consistency issues.
Recognizing this challenge, recent approaches~\cite{zeng24paint3d, siddiqui2024assetgen, deng2024flashtex, kim2024rocotex} often focus on synchronized output within a limited number of views, typically 4 or 6 orthogonal viewpoints, to improve alignment throughout the entire model.

Despite these impressive results, the limited number of views inevitably yields occlusions, leading to incomplete textures. 
Aware of this problem, various strategies have been explored, especially for inpainting using ControlNet in UV space~\cite{zeng24paint3d, siddiqui2024assetgen}. 
However, these solutions frequently introduce locality biases in UV space, which do not accurately reflect the 3D once mapped onto a 2D Cartesian plane. 
As a result, such methods often produce degraded details in occluded areas, compromising texture coherence.

Concurrent work, MVPaint~\cite{24mvpaint}, has similarly identified these limitations, suggesting surface-based extrapolation as a potential solution. 
In contrast, we address this issue using 3D-aware occlusion inpainting, which preserves texture fidelity and enhances overall consistency.

\section{Methodology}
\label{sec:method}
\begin{figure*}[tp]
    \centering
    \includegraphics[width=\linewidth]{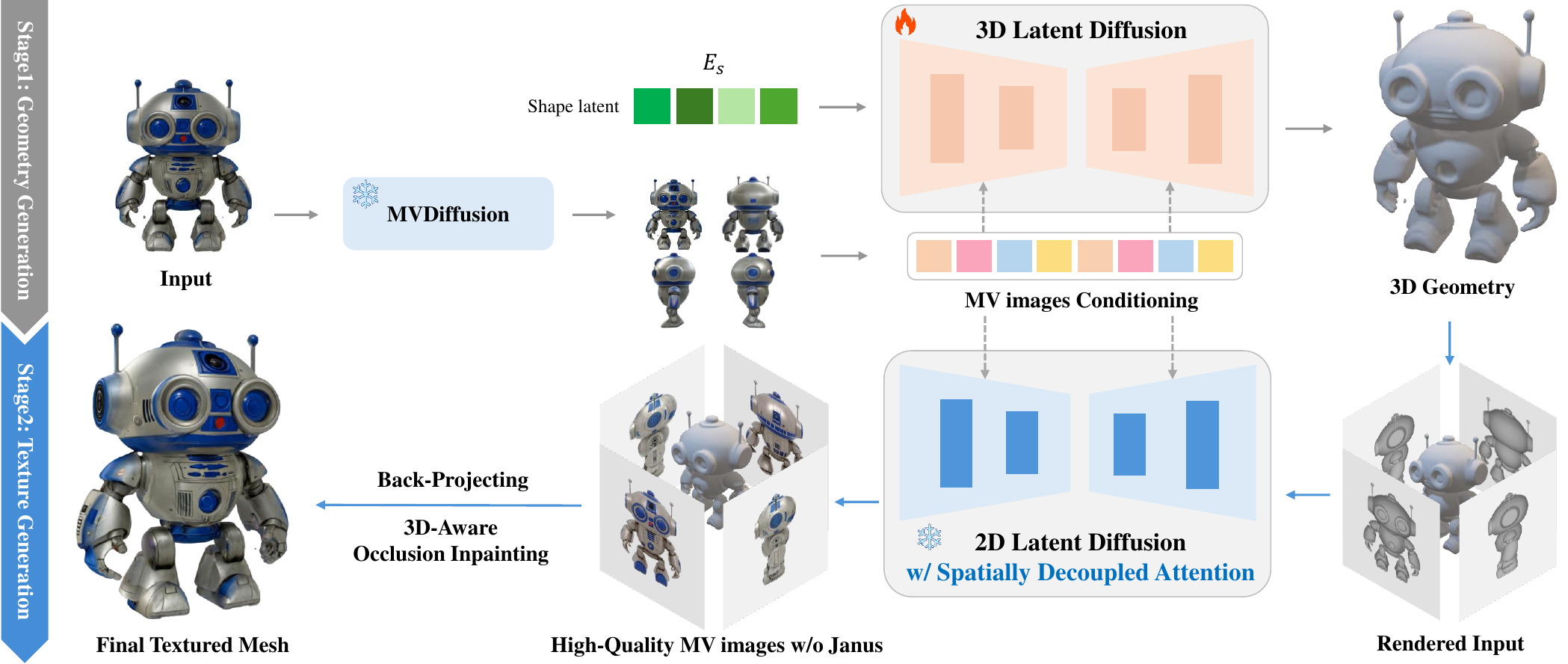}
    \caption{\textbf{CaPa pipeline}. 
     We first generate 3D geometry using a 3D latent diffusion model. 
     Using the learned 3D latent space with ShapeVAE, we train a 3D Latent Diffusion Model that generates 3D geometries, guided by multi-view images to ensure alignment between the generated shape and texture. 
     After the 3D geometry is created, we render four orthogonal views of the mesh, which serve as inputs for texture generation. 
     To produce a high-quality texture while preventing the Janus problem, we utilize a novel, model-agnostic spatially decoupled attention. 
     Finally, we obtain a hyper-quality textured mesh through back projection and a 3D-aware occlusion inpainting algorithm.
    }
    \label{fig:pipeline}
\end{figure*}

This section introduces CaPa, a novel 3D asset generation framework. 
CaPa operates in two stages: (1) 3D geometry synthesis using multi-view guided 3D latent diffusion and (2) 2D texture generation via 2D latent diffusion. 
First, we generate a 3D geometry by leveraging multi-view guided 3D latent diffusion to capture structural details from diverse perspectives. 
Second, to create an initial texture, we synthesize four orthogonal view images along the generated mesh. 
The untextured regions, caused by occlusions, are then completed using a 3D-aware inpainting algorithm to ensure a seamless appearance across all surfaces. 
Both stages are aligned through shared multi-view images, ensuring consistent geometry and texture. 
Using a fully feed-forward approach, the entire 3D asset generation process is completed in less than 30 seconds. 
Figure~\ref{fig:pipeline} illustrates the CaPa pipeline.

\subsection{Geometry Generation via 3D Latent Diffusion}

For 3D geometry generation, we leverage the recently proposed ShapeVAE to model the distribution of 3D geometry, while ensuring alignment with both image and text latent spaces. Using ShapeVAE, we train a 3D latent diffusion model~\cite{rombach22ldm} guided by multi-view (MV) images to achieve better alignment with texture generation. 
This MV-conditioned 3D latent diffusion model can efficiently generate high-quality 3D shapes, even under complex scenarios.

\subsubsection{Latent Space for Geometry Representation} 
We adopt a perceiver-based encoder that captures essential 3D geometric features, in line with Michelangelo~\cite{zhao23Michelangelo}. 
The encoder is trained to reconstruct neural fields from input point clouds, ensuring efficient representation learning.
\newline

\noindent{\textbf{Shape Encoder.}}
Given a ground-truth 3D shape, we aim to encode it into a latent representation, $\mathbf{E}_s$. First, we sample points from the surface $\mathbf{P} \in \mathbb{R}^{N \times 3}$, along with their corresponding normal vectors $\mathbf{n}$. These points and normals are then processed using cross-attention mechanisms that incorporate Fourier encodings and spherical harmonics to capture fine geometric details. 
\newline

\noindent{\textbf{Shape Decoder.}}
The perceiver-based decoder $\mathcal{D}$ reconstructs the neural field from latent embeddings $\mathbf{E}_s$. Given a query point $\mathbf{p} \in \mathbb{R}^3$ and $\mathbf{E}_s$, the decoder predicts the occupancy value for each point. The reconstruction process is optimized using binary cross-entropy (BCE) loss, 
\begin{equation}
\label{eq:vae_loss}
\mathcal{L}_{\text{vae}} = \mathbb{E}_{\mathbf{p} \in \mathbb{R}^3} \left[\text{BCE}\left(\hat{\mathcal{O}}(\mathbf{p}), \mathcal{D}\left(\mathbf{p} | \mathbf{E}_s \right)\right)\right] + \lambda_{\text{kl}} \mathcal{L}_{\text{kl}}
\end{equation}
where $\hat{\mathcal{O}}(\mathbf{p})$ represents the occupancy prediction for the query point $\mathbf{p}$, and $\mathcal{L}_\text{kl}$ for KL divergence regularization.

\subsubsection{Multi-View Guided 3D Latent Diffusion}
Leveraging the learned latent space, we employ a 3D latent diffusion model to generate geometry. 
However, due to its reliance on the latent space learned by ShapeVAE, this model often demonstrates limited geometric understanding. 
Moreover, establishing a unified generation pipeline requires consistency between the generated shape and texture.

To address these challenges, we train a multi-view guided 3D latent diffusion model that incorporates richer geometric priors and maintains alignment between shape and texture. Specifically, we use an off-the-shelf MV diffusion model to generate a set of multi-view images $\hat{\mathbf{I}}$ from the input condition (\textit{e.g.,} an image or text). 
This guidance provides stronger priors for 3D shape generation, ensuring improved alignment with texture synthesis in subsequent steps.

Finally, we train the multi-view guided 3D diffusion model $\epsilon_\theta$. 
The training objective follows the standard reverse denoising process of latent diffusion~\cite{rombach22ldm, ho20ddpm, song2021ddim}:
\begin{equation}
\mathcal{L}_{\text{3D-LDM}} := \mathbb{E}_{\epsilon \sim \mathcal{N}(0, 1), t} \left[\|\epsilon - \epsilon_\theta(\mathbf{E}_{s}, t, \tau_\theta(\hat{\mathbf{I}}, \pi))\|_2^2\right]
\end{equation}
where $\tau_\theta$ extracts camera embeddings from the multi-view images $\hat{\mathbf{I}}$ to incorporate camera-specific parameters $\pi_i$.

After applying the Marching Cubes~\cite{marchingcube} to extract a 3D mesh from the occupancy field, we employ a robust remeshing process to ensure smooth and high-quality manifold geometry. 
We carefully design the post-processing pipeline to address common meshing challenges, such as aliasing and non-manifold structures.
This remeshing strategy also supports artist-preferred quadrilateral meshes and high-resolution UV-mapped textures. 
For a comprehensive discussion, please refer to Appendix~\ref{app:mesh}.
\subsection{Texture Generation for Input Geometry}
For texture generation, we first render four orthogonal views of the generated mesh, which serve as the input for texture synthesis.
Ensuring texture consistency across these multiple views is critical but inherently challenging, as traditional approaches often struggle with the Janus problem.

A conventional solution for multi-view texture generation involves using MVDiffusions with geometry-guidance ControlNet~\cite{zhang23controlnet}, 
as suggested in CLAY~\cite{Zhang24clay} and MVPaint~\cite{24mvpaint}.
However, these approaches encounter several limitations: 
the output resolution for each view in MVDiffusion models typically remains constrained to a low 256-pixel resolution, resulting in suboptimal visual fidelity. 
Additionally, existing MVDiffusions rely on modified U-Net~\cite{Ronneberger15unet} architectures that differ from the original 2D latent diffusion model~\cite{rombach22ldm}, thus precluding the reuse of pre-trained ControlNet and necessitating resource-intensive retraining.

\subsubsection{Spatially Decoupled Cross Attention}
In response to these challenges, we introduce the \textit{Spatially Decoupled Cross Attention}, 
a model-agnostic approach that amplifies multi-view information.
To ensure texture consistency, we first concatenate the rendered images into a single batch and synthesize textures for all views simultaneously. 
Feature channels are then replicated and assigned to view-specific guidance, allowing each channel to focus on its respective spatial region. 
\newline

\begin{figure}[tp]
    \centering
    \includegraphics[width=\linewidth]{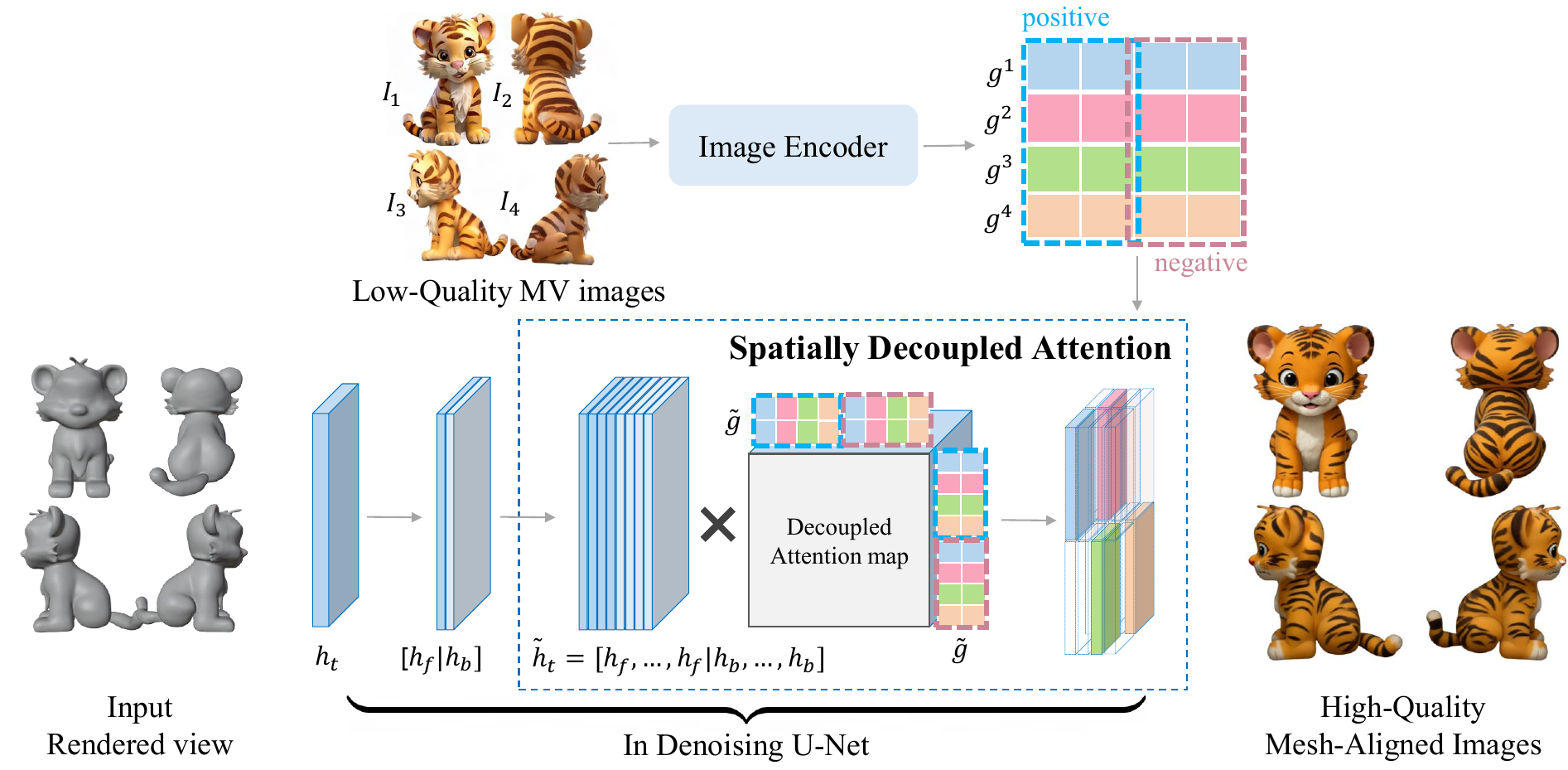}
    \caption{\textbf{Spatially Decoupled Cross Attention}. 
    To produce high-quality multi-view images for a given geometry, we design a model-agnostic Spatially Decoupled Cross Attention. 
    During cross-attention in denoising U-Net, we replicate hidden feature channels so that each duplicated channels focuses solely on the designated view.
    Since the design is model-agnostic, we can utilize an external ControlNet to guide the textures aligned with the input mesh. 
    }
    \label{fig:regional_ip_adapter}
\end{figure}

\noindent{\textbf{View-Specific Feature Enhancement.}}
For a hidden feature $h_t \in \mathbb{R}^{C \times W \times H}$ at time step $t$, we split the feature channels into positive $h_f$ and negative components $h_b$, both of size $h_f, h_b \in \mathbb{R}^{C/2 \times W \times H}$. 
These features are then replicated across the $N$ views: 
\begin{equation} \label{eq:hidden_feature
} \tilde{h}_t = \big[\underbrace{h_f, \dots, h_f}_{N} \mid \underbrace{h_b, \dots, h_b}_{N}\big] \in \mathbb{R}^{CN \times W \times H}.
\end{equation} 
Then, we independently encode each low-quality multi-view guide image and combine them into a unified feature vector:
\begin{equation} 
\label{eq:image_feature} 
\tilde{g} = \left[{g^1_f, \dots, g^N_f} | {g^1_b, \dots, g^N_b} \right] \in \mathbb{R}^{CN \times F} .
\end{equation}
The combined feature $\tilde{g}$, together with replicated hidden states $\tilde{h}_t$, undergoes cross-attention, 
\begin{equation} 
\label{eq:decoupled_cross_attention} 
Z = \text{XAttn}(\tilde{g}, \tilde{h}_t)
\end{equation}
As a result of the attention, each channel group focuses on its designated view, preserving view-specific information. 
Finally, we aggregate the features corresponding only to the spatial region of the $i$-th view, ensuring that the guidance is attended to only in its respective region.
Figure~\ref{fig:regional_ip_adapter} illustrates the spatially decoupled attention mechanism.

\noindent{\textbf{Advantages of Spatially Decoupled Attention.}}
The proposed mechanism allows each view’s spatial region to be processed independently within the same U-Net architecture, preserving view-specific details and ensuring consistency across perspectives.
Therefore, it can utilize large pre-trained generative models like ControlNet or SDXL~\cite{podell2024sdxl} without any architectural modifications or finetuning. 

Intuitively, this approach \textit{amplifies} low-quality, geometry-unaligned multi-view guide images to high-quality, geometry-aligned multi-view images in a single diffusion process.
Unlike conventional MVDiffusions, which require modified U-Net architecture and extensive retraining, the proposed method generates consistent, high-fidelity textures in a model-agnostic manner, bypassing traditional limitations.

\subsubsection{Occlusion Inpainting} 
In the initial texture generation, we create four orthogonal images for 3D geometry, which are then back-projected to produce a coarse-textured mesh. 
However, due to the limited number of views, occlusions inevitably arise in regions not visible from these views.
To address this issue, we propose a novel 3D-aware occlusion inpainting algorithm.
\newline

\begin{figure}[tp]
    \centering
    \includegraphics[width=\linewidth]{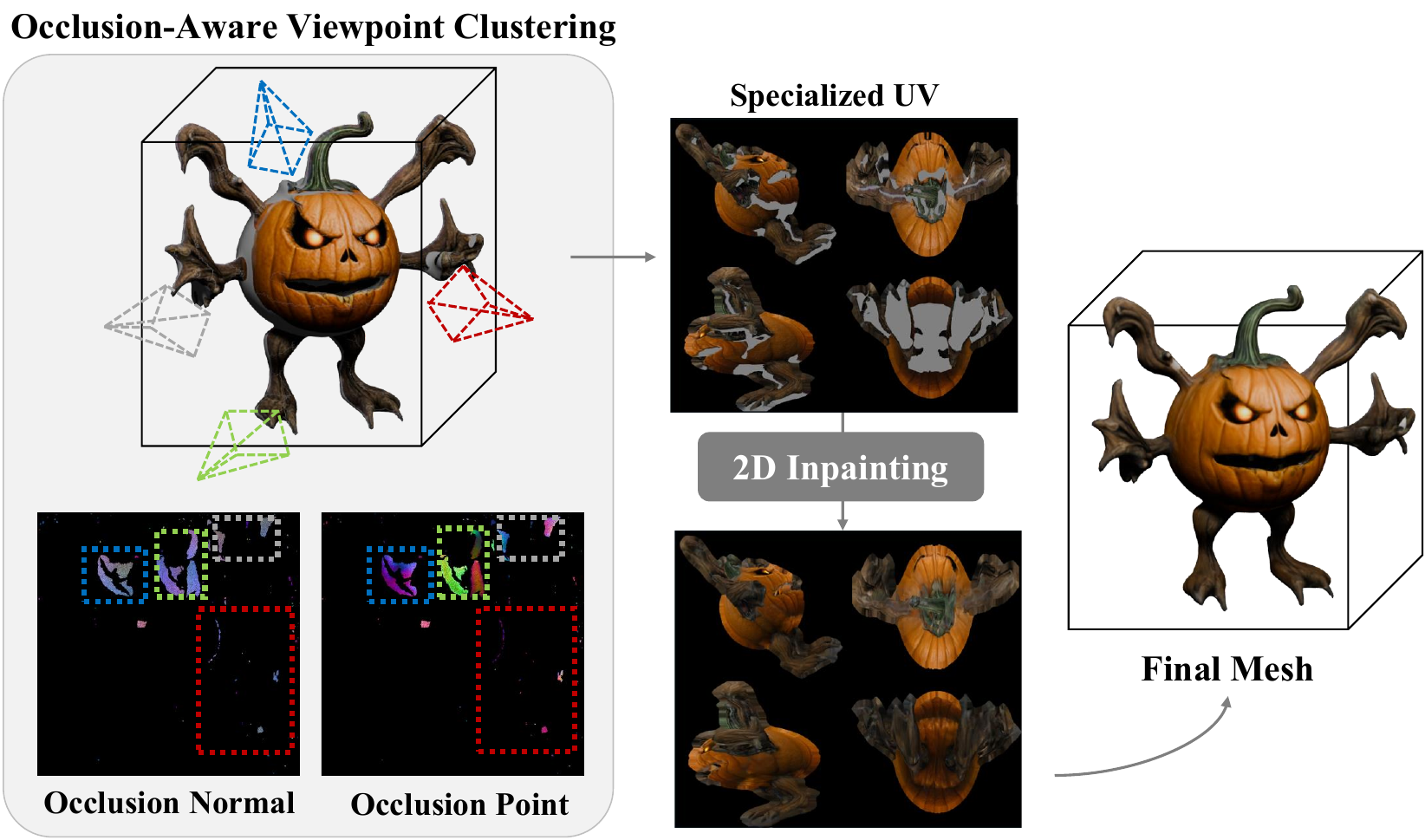}
    \caption{\textbf{3D-Aware Occlusion Inpainting.} 
    First, we cluster the normal and spatial coordinates of the occluded face. 
    Using clustered centers as viewpoints, we create specialized UV maps through projection mapping. 
    This approach captures surface locality, allowing 2D diffusion-based inpainting to effectively fill occluded regions. Note that this UV map is utilized solely for occlusion.
    }
    \label{fig:occlusion_inpainting}
\end{figure}

\noindent{\textbf{3D-Aware Occlusion Mapping and Inpainting.}}
As discussed in Section~\ref{sec2.3}, existing occlusion inpainting methods have challenges for speed or locality bias. 
To overcome these issues, we propose a more adaptive, geometry-aware occlusion inpainting approach. 

This method identifies and isolates the untextured regions on the mesh surface, targeting only the areas affected by occlusion. 
To capture the occluded areas more accurately, we employ $k$-means clustering using the face normal orientations and spatial coordinates of untextured faces on the 3D mesh surface. 
This clustering process yields several representative cluster centroids, each corresponding to a distinct occluded region on the mesh. 
For each centroid, we define a viewpoint on the unit sphere that faces the untextured area, projecting the coarse-textured mesh from this viewpoint to capture a region-specific view of the occlusion.

We then aggregate these projections onto a single 2D plane, creating a \textit{specialized UV map} dedicated solely to inpainting occluded areas. 
Unlike traditional UV maps, this occlusion-specific UV map maintains a more accurate representation of locality on the mesh surface, thus preserving the spatial coherence required for 2D inpainting methods.  
By leveraging this UV map, we use the same 2D diffusion model for texture generation and inpainting. 
This approach yields a high-fidelity, 3D-aware texture that effectively fills occluded regions. 
Empirically, we found that $k=6$ covers most occluded areas, and further apply the extrapolation technique for other remaining regions.
Figure~\ref{fig:occlusion_inpainting} illustrates the 3D-aware occlusion inpainting scheme.

\section{Experiments}
\label{sec:experiments}
\subsection{Implementation Details}
\noindent{\textbf{Dataset.}}
We utilize the Objaverse~\cite{objaverse} dataset for ShapeVAE and 3D latent diffusion training. 
We preprocess this dataset by filtering out low-quality, non-manifold meshes, retaining 150K high-quality objects with high CLIP~\cite{radford21clip} scores. 
\newline

\noindent{\textbf{Geometry Generation.}}
We employ a perceiver-based~\cite{perceiver} transformer architecture for ShapeVAE, as used in Michelangelo~\cite{zhao23Michelangelo}, modifying it with Fourier featuring and spherical harmonics encoding. 
For the multi-view guided 3D latent diffusion stage, we adopt a U-Net-based transformer. 
The entire training takes around 8 days on 32 NVIDIA A100 GPUs.
Finally, we integrate a custom-trained MVDiffusion model, following the setup in ImageDream~\cite{wang2023imagedream}, but modify it to treat the input view as the frontal view rather than estimating.
\newline

\noindent{\textbf{Texture Generation.}}
We implement the entire texture generation mechanism for texture synthesis upon the SDXL~\cite{podell2024sdxl}. 
We utilize pre-trained IP Adapter~\cite{ye2023ip-adapter} layer and CLIP~\cite{radford21clip} for the image feature extractor, Depth-ControlNet~\cite{zhang23controlnet} for a geometric guide. 
Appendix~\ref{app:texture_synthesis} provides the additional implementation specifications.

\subsection{Qualitative Comparison}
\subsubsection{Comparison with State-of-the-Art Methods}
We evaluate the proposed CaPa by comparing it with various state-of-the-art image-to-3D approaches. 
Specifically, we include SDS-based~\cite{poole2023dreamfusion} technique: DreamCraft3D~\cite{sun2024dreamcraftd}; 
MVDiffusion with its correponding normals for sparse neural reconstruction:  Era3D~\cite{li2024era3dhighresolutionmultiviewdiffusion} (an enhanced version of Wonder3D~\cite{long2024wonder3d}), or Unique3D~\cite{wu2024unique3d} for mesh optimization; and finally, the LRM-based method: SF3D~\cite{boss2024sf3d}.
Note that all assets were converted into a polygonal mesh using official codes to ensure consistency and fair evaluation. 
\newline

\begin{figure*}[tp]
    \centering
    \includegraphics[width=0.9\linewidth]{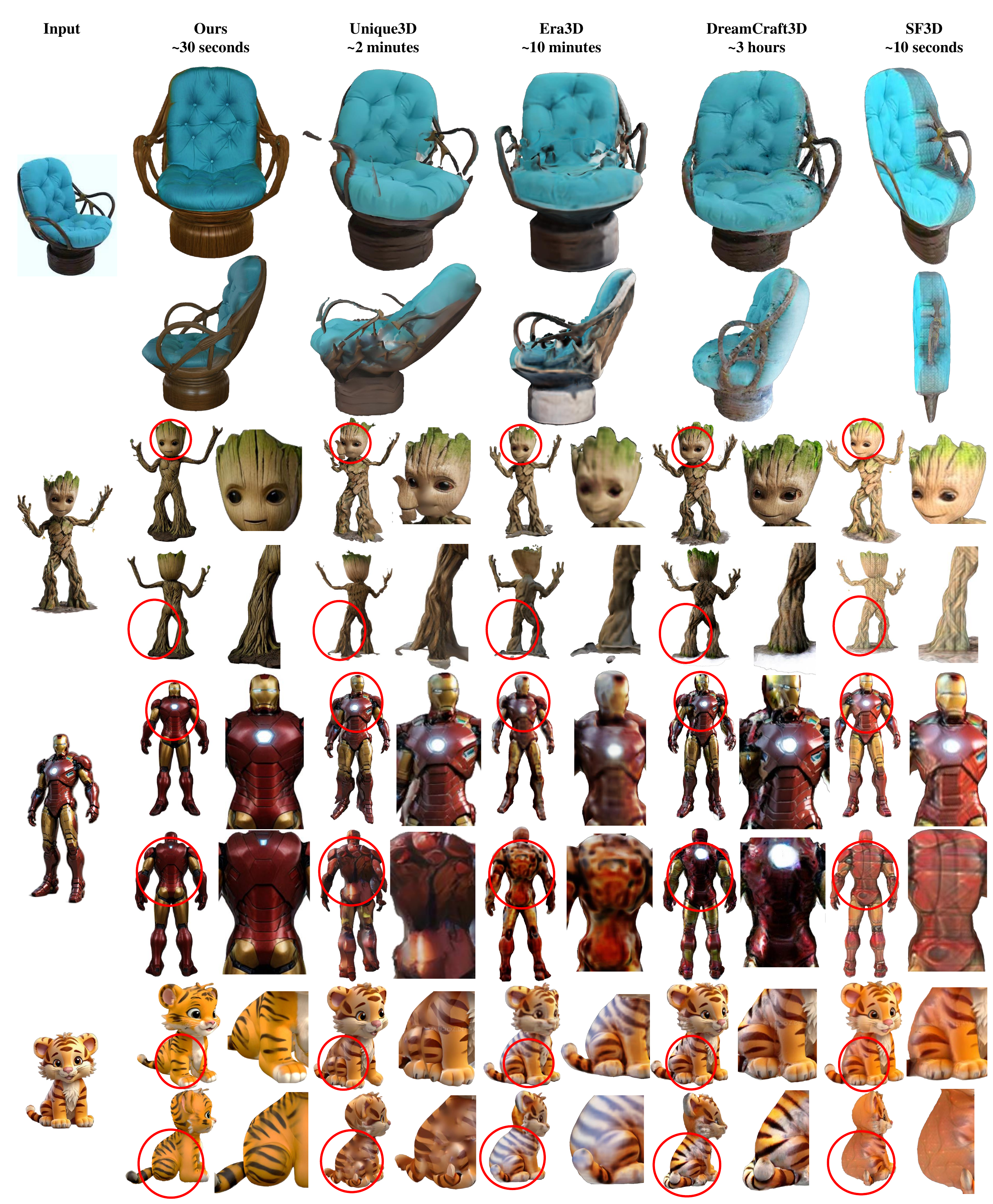}
    \caption{\textbf{Qualitative Comparison of Image-to-3D Generation}. 
    We compare CaPa with state-of-the-art Image-to-3D methods. 
    Here, all the assets are converted to \textit{polygonal mesh}, using its official code. 
    The proposed CaPa significantly outperforms both geometry stability and texture quality, especially for the back and side view's visual fidelity and texture coherence. 
    }
    \label{fig:main_qual}
\end{figure*}

\noindent{\textbf{High-Fidelity Mesh with Consistent Multi-View Textures.}}
As shown in Figure~\ref{fig:main_qual}, CaPa significantly outperforms existing state-of-the-art methods, especially in not just frontal view, but also in back and side views, where competitors typically exhibit substantial degradation. 
Compared to sparse view NeRF reconstruction methods like Era3D~\cite{li2024era3dhighresolutionmultiviewdiffusion}, CaPa demonstrates markedly superior texture quality. 
In relation to Unique3D~\cite{wu2024unique3d}, which generates assets through mesh optimization based on multi-view images and corresponding normals, CaPa exhibits significantly higher geometric stability (See the second asset). 
Finally, compared to the SDS-based method~\cite{sun2024dreamcraftd}, CaPa consistently produces high-fidelity polygonal meshes while accelerating the generation process and improving texture quality across multiple views. 
These results highlight CaPa's effectiveness and robustness in producing high-quality 3D assets, showcasing its advantages over current methodologies in the field.

\begin{figure}[h]
    \centering
    \includegraphics[width=\linewidth]{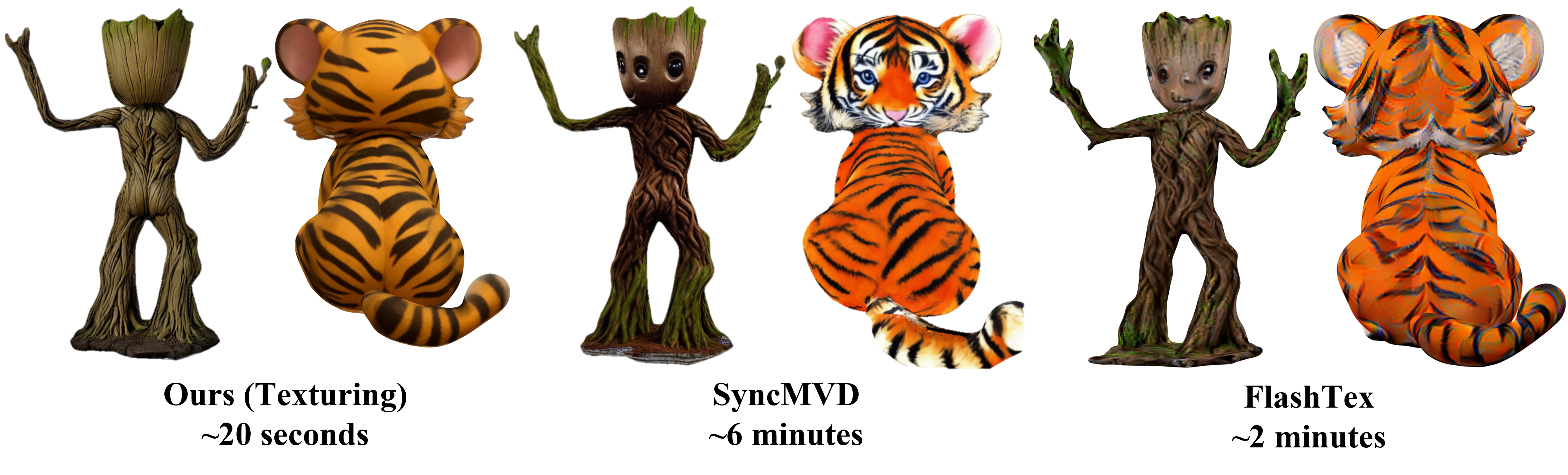}
    \caption{\textbf{Comparison of Texturing Method.} 
    Unlike prior works, CaPa effectively resolved the Janus problem with consistent ID.
    }
    \label{fig:texture}
\end{figure}

To further validate the effectiveness of CaPa's texture generation process, we compared the textured outputs of our mesh with results from SyncMVD~\cite{liu2023syncmvd} and FlashTex~\cite{deng2024flashtex}, a leading texturing method. 
As shown in Figure~\ref{fig:texture}, CaPa successfully addresses the Janus problem. In contrast, previous methods demonstrate limited id-consistency and still exhibit the Janus problem, relying primarily on \textit{textual} input alone.

\subsection{Quantitative Comparison}

We present the quantitative results in Table~\ref{table:quan}, using 50 randomly selected test images generated via GPT~\cite{openai2024gpt4technicalreport}. 
We report the CLIP~\cite{radford21clip} score and FID~\cite{17fid}, evaluated by measuring the similarity between each input image and the corresponding rendered outputs across 30 random views, thereby indirectly assessing multi-view consistency. 
As shown in the table, CaPa achieves significantly higher scores than SOTA techniques, demonstrating its capability to produce consistent textures while preserving alignment with the input.
\begin{table}[h]
    \centering
    \small
    \begin{tabular}{c c c c c}
         \toprule
         \multirow{1}{*}{{Method}} 
         &\multicolumn{1}{c}{CLIP ($\rm I$-$\rm I$) $\uparrow$}
         &\multicolumn{1}{c}{FID $\downarrow$}
         &\multicolumn{1}{c}{Time $\downarrow$}
         \\
         
         \midrule    
         
         Ours 
         & \textbf{86.34}
         & \textbf{47.56}
         & \underbar{$\sim$30 seconds}
         \\
         
         DreamCraft3D~\cite{sun2024dreamcraftd}
         & 77.61
         & 75.66
         & $\sim$3 hours 
         \\

         Unique3D~\cite{wu2024unique3d} 
         & \underbar{81.92}
         & \underbar{67.17}
         & $\sim$2 minutes
         \\

         Era3D~\cite{li2024era3dhighresolutionmultiviewdiffusion} 
         & 66.81
         & 89.18
         & $\sim$10 minutes
         \\

         SF3D~\cite{boss2024sf3d} 
         & 70.18
         & 84.52
         & \textbf{$\sim$10 seconds}
         \\

         \bottomrule
    \end{tabular}
    \caption{
    \textbf{Quantitative results.  
    }
    CaPa outperforms all the competitors by a significant margin in both CLIP score and FID score, with a reasonable generation time.
    }
    \label{table:quan}
\end{table}
\vspace{-1em}

\subsection{Ablation Study and Discussion}
\begin{figure*}[tp]
    \centering
    \includegraphics[width=\linewidth]{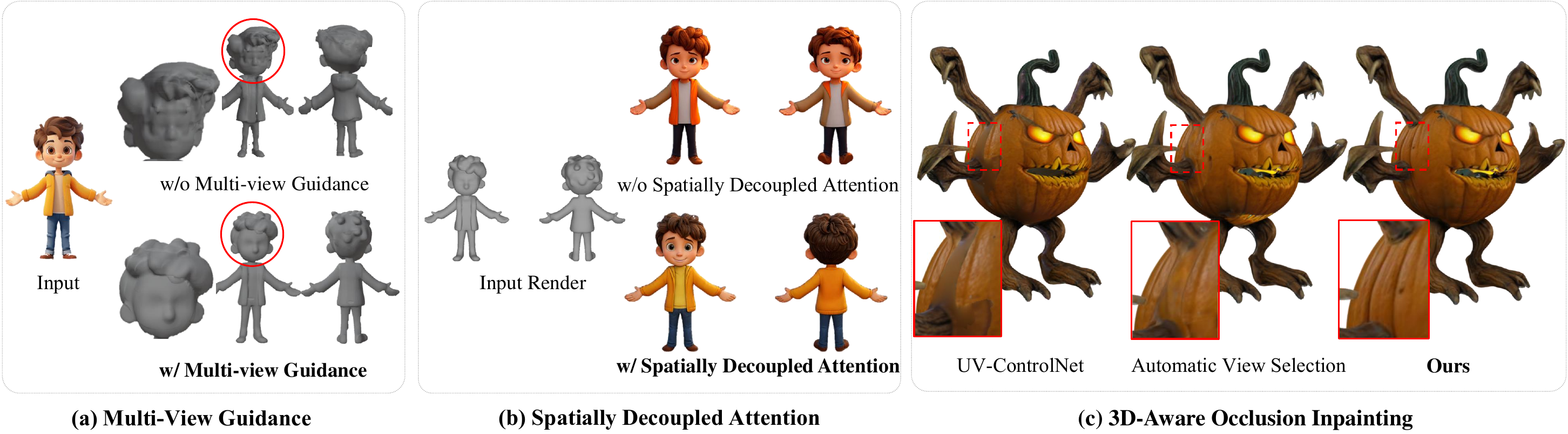}
    \caption{\textbf{Ablation Study.}
    (a) demonstrates that using multi-view guidance significantly increases the geometry quality. 
    (b) shows our Spatially Decoupled Attention effectively resolves the Janus problem, achieving high-fidelity texture coherence,
    (c) reveals our occlusion inpainting outperforms previous inpainting methods like UV-ControlNet, presented in Paint3D~\cite{zeng24paint3d}.
    }
    \label{fig:ablation}
\end{figure*}

We conduct comprehensive ablation studies to substantiate the effectiveness of each design element within CaPa, showing the importance of each component when the generation of high-quality 3D assets. 
Figure~\ref{fig:ablation} shows the ablation results of each component.
\newline

\noindent{\textbf{Multi-View Guidance in 3D Latent Diffusion.}}
As shown in Figure~\ref{fig:ablation} (a), the multi-view guidance for 3D latent diffusion provides more comprehensive information on 3D geometry, which is crucial for the stable generation of 3D assets. 
It enables smoother, higher-quality geometry generation and minimizes quality loss during the meshing.
\newline

\noindent{\textbf{Janus Prevention of Spatially Decoupled Attention.}}
The proposed Spatially Decoupled Attention effectively addresses the Janus problem by independently guiding each view during texture generation. 
Figure~\ref{fig:ablation} (b) shows the texture generation results of CaPa, both with and without Spatially Decoupled Attention, revealing the adapter's substantial capability to prevent multi-face generation. 
\newline

\noindent{\textbf{Occlusion Inpainting.}}
In this section, we demonstrate that our novel 3D-aware occlusion inpainting significantly outperforms previous occlusion inpainting solutions. 
In comparison to UV-ControlNet~\cite{zeng24paint3d} and automatic view-selection methods~\cite{chen23text2tex}, our approach achieves more comprehensive and seamless occlusion inpainting within a single diffusion inference step. This results in fewer visible seams and a more cohesive texture, as shown in Figure~\ref{fig:ablation} (c).

To further validate our inpainting method, we report the FID~\cite{17fid} and KID~\cite{18kid} scores between the generated textures and the occlusion-filled textures, as shown in Table~\ref{table:occlusion}. 
Lower FID and KID scores indicate better alignment and visual coherence. 
Our method achieves an FID of 55.23 and KID of 13.46, significantly outperforming UV-ControlNet, demonstrating its ability to preserve texture fidelity and semantic accuracy. 
While the automatic view-selection method~\cite{chen23text2tex} achieves slightly lower scores, it requires 20 seconds for iterative smoothing, resulting in blurred output. 
In contrast, our approach maintains sharpness and semantic fidelity while achieving faster, visually coherent textures.
\begin{table}[h]
    \centering
    \small
    \begin{tabular}{c c c c c}
         \toprule
         \multirow{1}{*}{{Method}} 
         &  \multicolumn{1}{c}{FID $\downarrow$}
         &  \multicolumn{1}{c}{KID $\downarrow$}
         &\multicolumn{1}{c}{Time $\downarrow$}
         \\
         
         \midrule    
         Ours 
         & \textbf{55.23}
         & \textbf{13.46}
         & \textbf{$\sim$5 sec.}
         \\

         Automatic View-Selection~\cite{chen23text2tex}
         & \underbar{62.31}
         & \underbar{15.83}
         & {$\sim$20 sec.}
         \\
         
         UV-ControlNet~\cite{zeng24paint3d}
         & 128.71
         & 37.38
         & \textbf{$\sim$5 sec.}
         \\
        
         \bottomrule
    \end{tabular}
    \caption{
    \textbf{Quantitative Comparison of Occlusion Inpainting.  
    }
    Our 3D-aware inpainting restores occlusions with minimal semantic drift and improves contextual alignment efficiently.
    }
    \label{table:occlusion}
\end{table}

\subsection{Scalability of CaPa}
\begin{figure}[h]
    \centering
    \includegraphics[width=\linewidth]{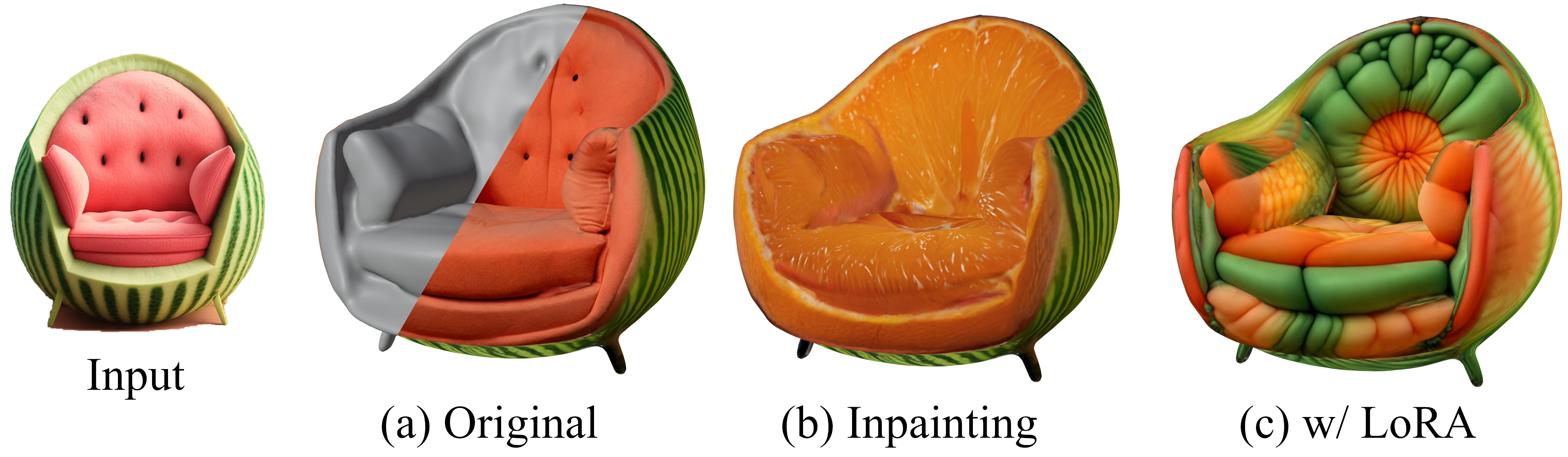}
    \caption{\textbf{Scalability of CaPa}.
    (a) Original result of CaPa. 
    (b) 3D inpainting result using text-prompt (``orange sofa, orange pulp''). CaPa's texture generation extends smoothly to 3D inpainting, stylizing the generated asset.
    (c) CaPa w/ LoRA~\cite{hu2022lora} adaptation. The model-agnostic approach allows CaPa to leverage pre-trained LoRA (balloon style) without additional 3D-specific retraining. 
    }
    \label{fig:lora}
\end{figure}

A major advantage of the Spatially Decoupled Attention in texture generation is its model-agnostic nature. 
Unlike previous MVDiffusion~\cite{shi2024mvdream, wang2023imagedream}, our texturing scheme can be directly applied to pre-existing large generative models and its variants, including LoRA~\cite{hu2022lora} and ControlNet. 
Additionally, the pipeline supports flexible extensions, such as 3D inpainting via text prompts.
We demonstrate CaPa's scalability through 3D inpainting and LoRA integration examples.
As shown in Figure~\ref{fig:lora}, CaPa requires no additional training for 3D stylization, underscoring its scalability with a wide range of 2D generative communities.
\newline

\subsection{Supplementary Material}
To further validate CaPa, we present additional results in Appendix~\ref{app:exps}, including diverse image/text-to-3D outputs, comparative experiments addressing Janus artifacts, and analysis of occlusion edge cases.
Additionally, we discuss the limitations of our work in Appendix~\ref{app:limit}. 
We kindly encourage readers to refer to these additional experiments.

\vspace{0.65em}
\section{Conclusion}
\label{sec:conclusion}
In this study, we propose CaPa, an efficient framework for high-quality 3D asset generation which separates 3D geometry from 2D texture synthesis. 
Using multi-view guided 3D latent diffusion of the occupancy field minimizes the quality loss during mesh extraction. 
For texture synthesis, a spatially decoupled cross-attention addresses the Janus problem without additional training. 
This model-agnostic solution integrates with large generative models like SDXL~\cite{podell2024sdxl} and achieves superior fidelity of the texture output. 
Finally, we present a novel 3D-aware occlusion inpainting algorithm, capturing 3D locality in UV space such that 2D diffusion-based inpainting effectively fills the occlusion.
All these stages are fully feed-forward, making the entire generation process complete in less than 30 seconds.
In summary, CaPa offers high-fidelity 3D synthesis at practical speeds, enabling immediate use in downstream applications. 

\clearpage
{\small
\bibliographystyle{ieeenat_fullname}
\bibliography{11_references}
}

\ifarxiv \clearpage \fi
\clearpage
\onecolumn
\appendix
\section*{Appendix}
\vspace{-0.5em}
\noindent\hrulefill
\vspace{-0.5em}
\section{Discussions and Limitations}
\label{app:limit}
In the main manuscript, we introduced CaPa, a carve-and-paint framework designed for fast, scalable, and high-fidelity 3D asset generation. While CaPa demonstrates significant practical advancements, some limitations remain.
\newline

\begin{figure}[h]
    \centering
    \includegraphics[width=0.8\linewidth]{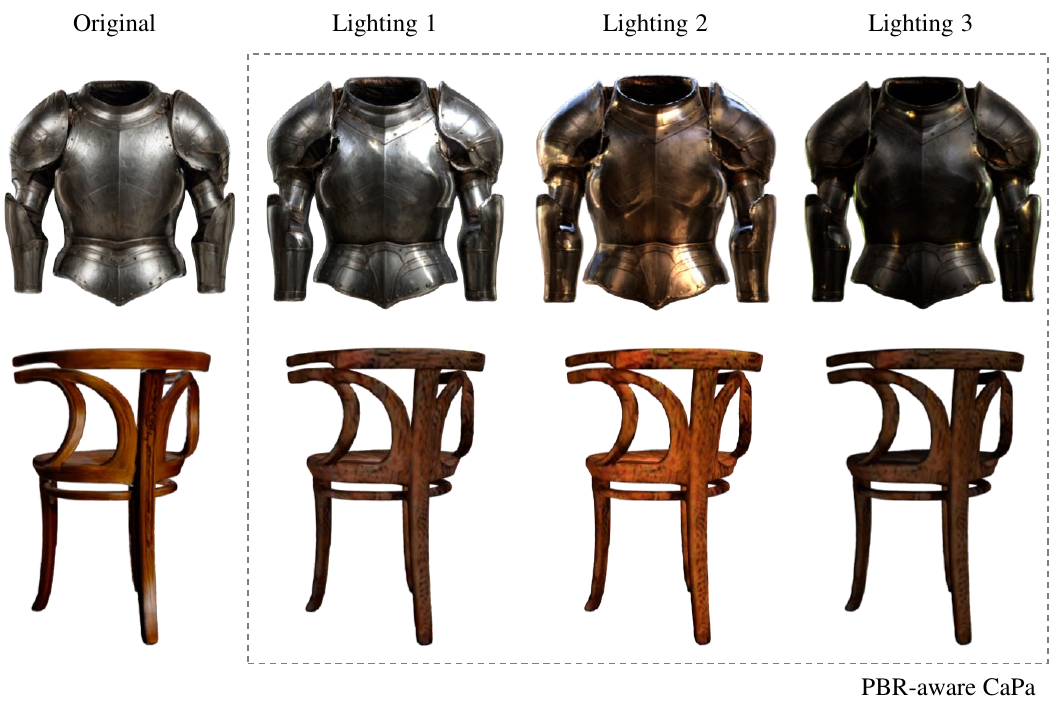}
    \caption{\textbf{Result of the CaPa with PBR Understanding.} 
    We demonstrate CaPa's capability for disentangling physically based rendering (PBR) materials. 
    The figure shows PBR-aware generation results under various lighting conditions: `city,' `studio,' and `night,' using Blender's default environment settings~\cite{Blender}.
    As shown, CaPa effectively adapts to different light environments, highlighting its potential for PBR-aware asset generation.
    }
    \label{fig:material}
\end{figure}

\noindent{\textbf{Understanding of Physically Based Rendering Material.}}
A notable limitation of the current CaPa pipeline is the lack of intrinsic support for Physically Based Rendering (PBR) materials.
While CaPa is designed to generate high-quality textures, it does not inherently address PBR material understanding.
Nonetheless, CaPa’s model-agnostic architecture allows for integration with external frameworks, making it compatible with recent material-aware diffusion models such as LightControlNet (as proposed in FlashTex~\cite{deng2024flashtex}) and MaterialFusion~\cite{litman2024materialfusion}. By leveraging these material-oriented models, CaPa can be extended to facilitate PBR-aware asset generation, as illustrated in Figure~\ref{fig:material}.

In line with FlashTex and MaterialFusion, we utilized SDS optimization to generate PBR-aware 3D assets with material diffusion models.
For additional information on SDS application for PBR separation, we refer readers to FlashTex~\cite{deng2024flashtex} and MaterialFusion~\cite{litman2024materialfusion}.
As demonstrated in Figure~\ref{fig:material}, CaPa exhibits adaptability in achieving PBR material understanding.
However, we observed that SDS optimization often led to a decrease in the original texture fidelity and increased overall generation time, typically extending beyond 10 minutes per asset. 
Additionally, we found that the process sometimes produced a diffuse appearance that differed slightly from the originally generated texture.

Future work will focus on balancing material disentanglement with CaPa’s original objective of rapid generation, aiming to achieve efficient, high-quality, and PBR-compatible textured mesh generation.
Building on CaPa’s base mesh and existing normal maps, one potential approach could involve training a diffusion model that takes normal map inputs to disentangle and separately generate diffuse, specular, and roughness maps. This would enable finer control over material properties, allowing CaPa to produce even more physically accurate textures while maintaining efficiency.
\begin{figure*}[h]
    \centering
    \includegraphics[width=\linewidth]{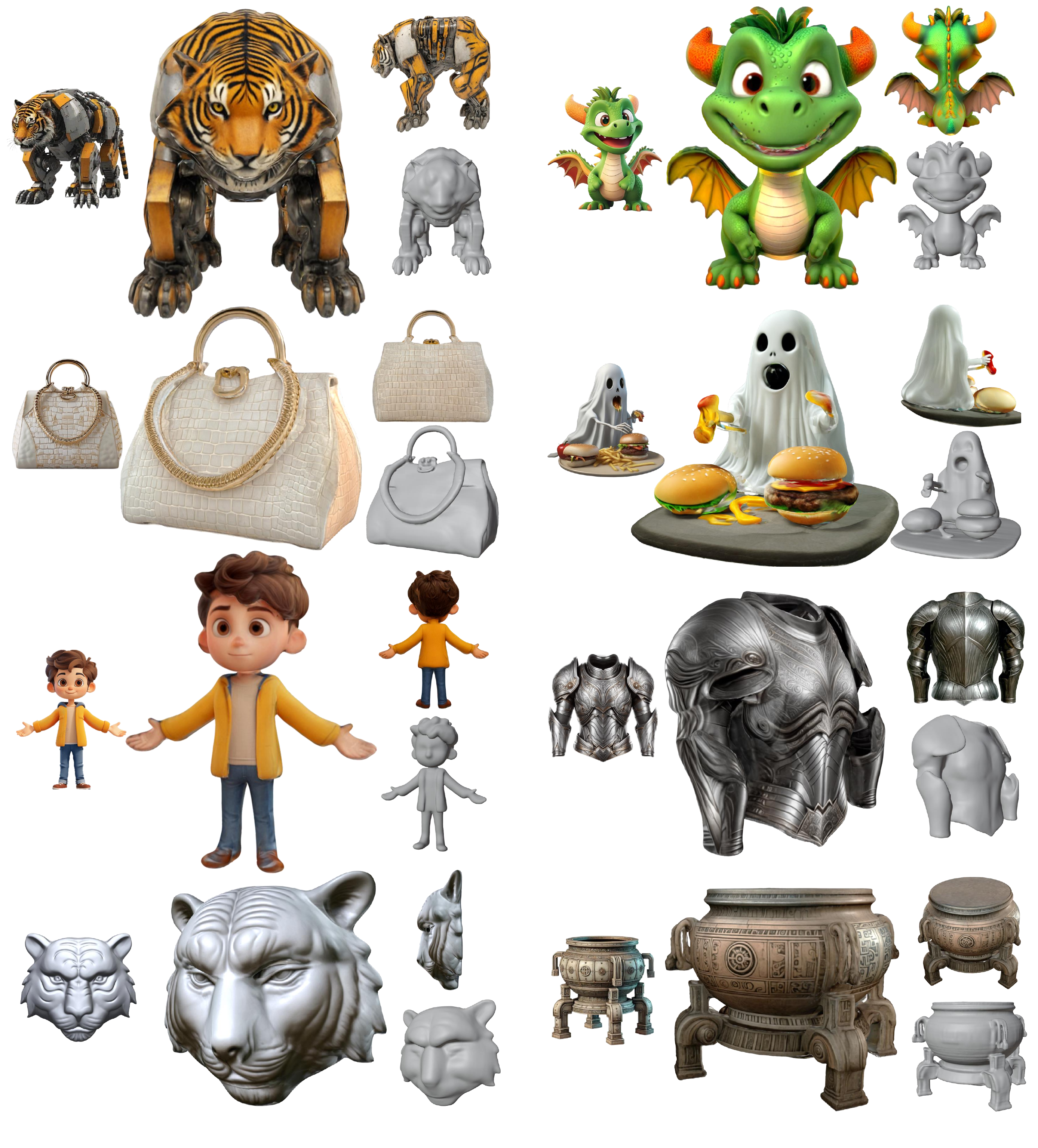}
    \caption{\textbf{Additional Image-to-3D Results of CaPa}.
    CaPa can generate diverse objects from textual, and visual input. 
    The result demonstrates our diversity across the various categories, marking a significant advancement in practical 3D asset generation methodologies. 
    }
    \label{fig:image1}
\end{figure*}
\clearpage
\begin{figure*}[h]
    \centering
    \includegraphics[width=0.95\linewidth]{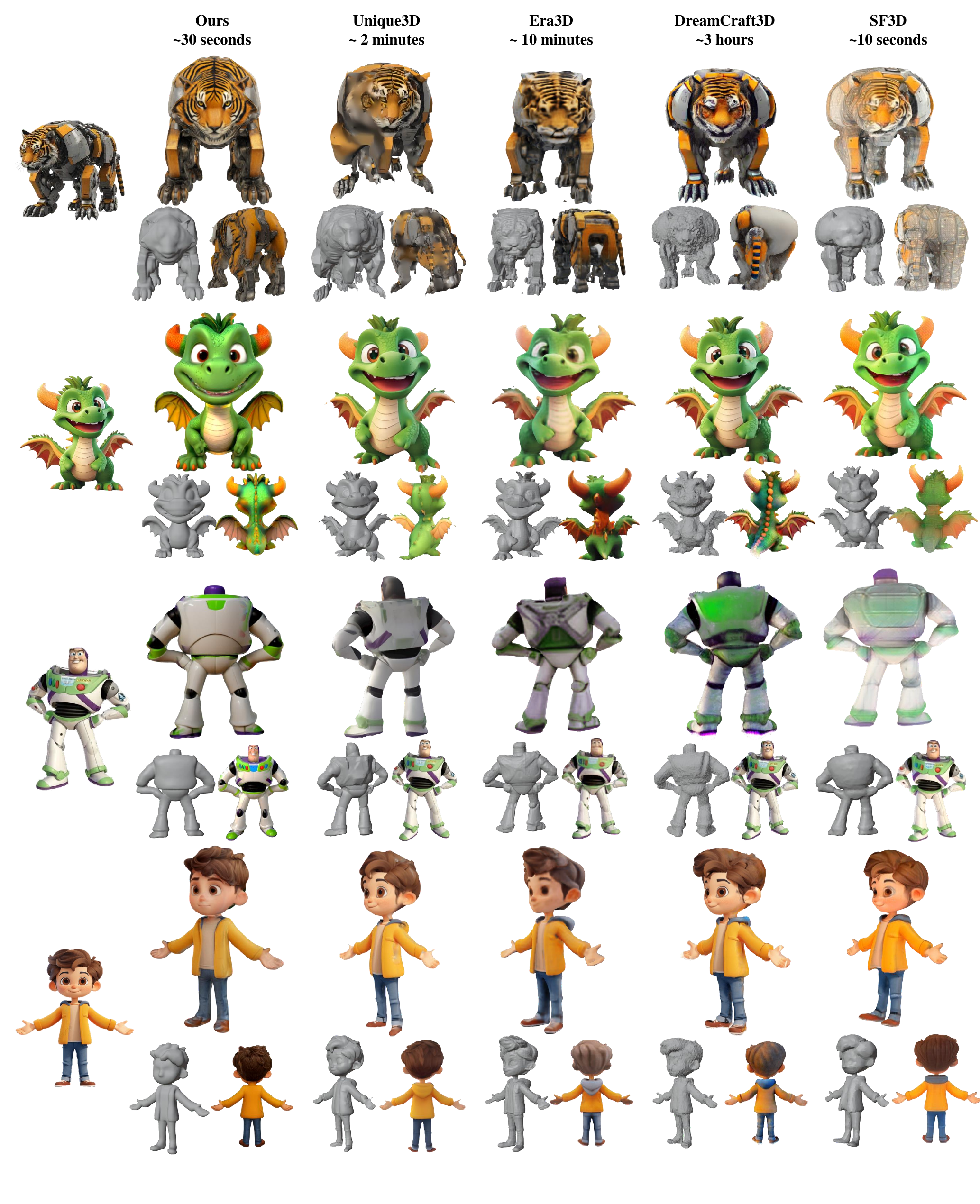}
    \caption{\textbf{Additional Comparison of Image-to-3D Generation}.
    CaPa significantly outperforms both geometry stability and texture quality, especially for the back and side view’s visual fidelity and texture coherence.
    }
    \label{fig:image2}
\end{figure*}
\section{Additional Experiments}
\label{app:exps}
\subsection{Additional Image-to-3D Results of CaPa}
In Figure~\ref{fig:image1}, we present additional image-to-3D asset generation results, with close-up renderings highlighting CaPa’s superior texture and geometric stability fidelity. 
Each row illustrates a detailed comparison between the base geometry and the final textured model, highlighting CaPa’s ability to produce cohesive and visually rich textures. 
Close-up views underscore the fidelity of fine details, such as the texture of the tiger’s skin, dragon's scales, and intricate patterns on the bag.
As shown, CaPa achieves state-of-the-art quality in texture coherence and structural consistency.

We present additional comparative experiments of state-of-the-art Image-to-3D asset generation methods in Figure~\ref{fig:image2}.
Consistent with Section~\ref{sec:experiments} in the main manuscript, we validate CaPa against  Unique3D~\cite{wu2024unique3d}, Era3D~\cite{li2024era3dhighresolutionmultiviewdiffusion}, DreamCraft3D~\cite{sun2024dreamcraftd}, and SF3D~\cite{boss2024sf3d}. 
Each row represents a different 3D asset (\textit{e.g.,} tiger, dragon, toy robot, and character model) with full-color renderings and corresponding grayscale meshes for each method.
As shown in Figure~\ref{fig:image2}, CaPa consistently delivers high-quality textures and detailed geometry within a significantly reduced runtime (around 30 seconds) compared to other methods.
This comparison highlights the efficiency and fidelity balance achieved by our approach.

\subsection{Impact of Spatially Decoupled Cross Attention on Janus Artifacts}
\subsubsection{Qualitative Analysis}
In this section, we present a qualitative comparison to demonstrate the effectiveness of the proposed spatially decoupled cross-attention mechanism in mitigating the Janus problem, a common issue in 3D asset generation when handling multi-view inputs.
\begin{figure*}[b]
    \centering
    \includegraphics[width=0.75\linewidth]{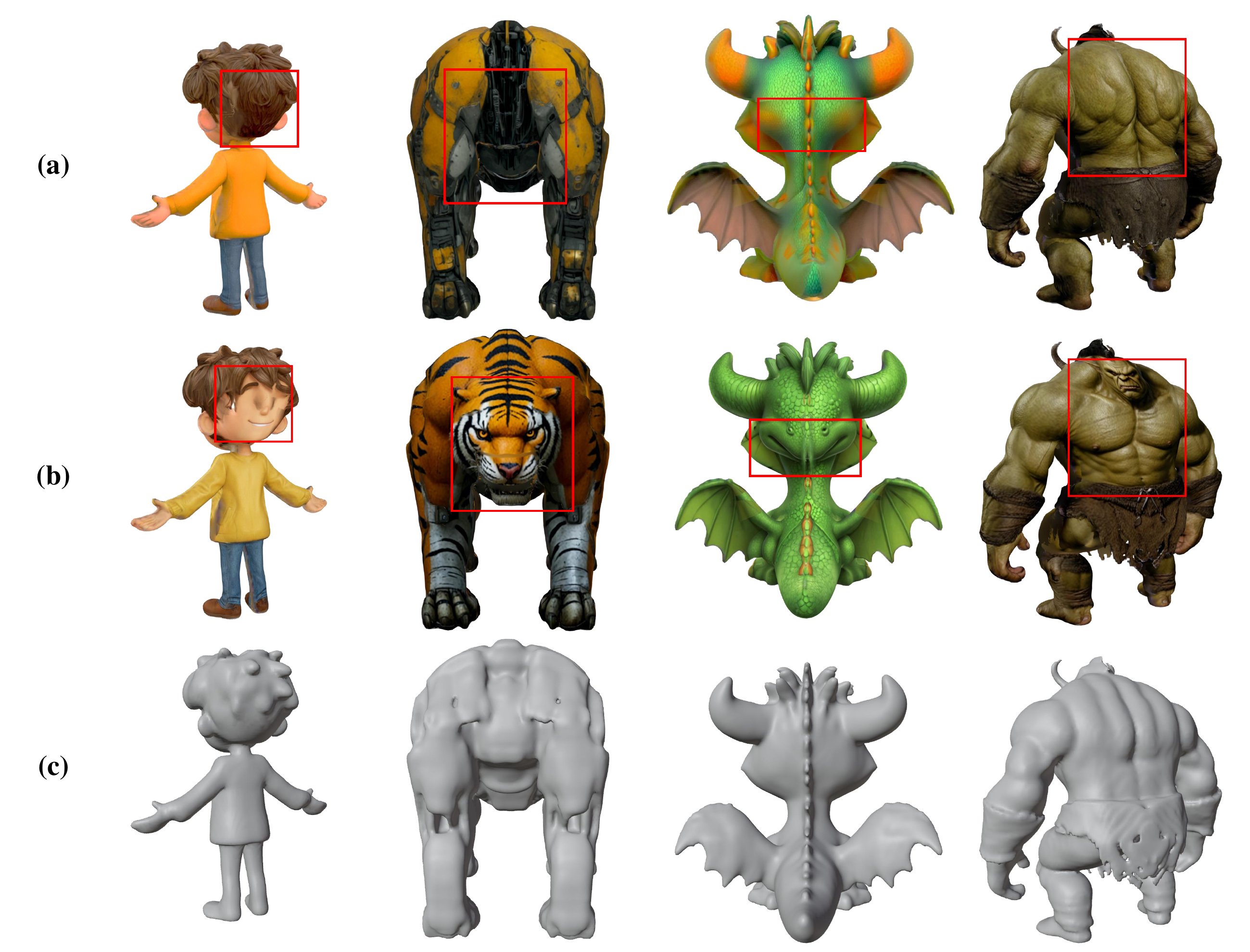}
    \caption{\textbf{Impact of Spatially Decoupled Cross Attention on Janus Artifacts}.
    In this additional figure, We demonstrate the capability of Janus prevention in the proposed spatially decoupled cross-attention mechanism.
    Each row depicts, (a) with spatially decoupled cross attention, (b) without spatially decoupled cross attention, and (c) a mesh rendering of the current view, respectively.
    }
    \label{fig:janus_prevention}
\end{figure*}
As seen in Figure~\ref{fig:janus_prevention}, spatially decoupled cross attention markedly improves multi-view consistency and eliminates Janus artifacts, demonstrating the necessity of spatial decoupling when incorporating multi-view guidance. 
This mechanism thus enhances CaPa's ability to generate high-quality 3D assets with accurate textures and structure across all views, supporting its robustness for diverse applications.

\subsubsection{Quantitative Analysis}
Additionally, we present the quantitative analysis of the Janus prevention capability of the CaPa. 
Before the discussion, we acknowledge that directly evaluating multi-view inconsistency is challenging, as traditional metrics often do not capture the specific artifacts introduced by the Janus problem.
To circumvent this problem, we assessed the impact of the Janus effect by measuring CLIP similarity between rendered normal maps and texture.
We hypothesize that the presence of Janus artifacts leads to a misalignment between the geometry and the texture's appearance, resulting in a decrease in the CLIP similarity score. 
Using this assumption, CLIP similarity serves as a proxy metric to evaluate the Janus artifacts assessment.

As shown in Table~\ref{table:janus}, with spatially decoupled attention, the average similarity score was 85.37, compared to 81.28 without it. Although the difference is modest, likely due to initial geometry guidance aligning texture and mesh. 
This result supports that spatially decoupled attention resolves the Janus effect and improves multi-view consistency in generated 3D assets.
\begin{table}[h]
    \centering
    \small
    \begin{tabular}{c c}
         \toprule
         \multirow{1}{*}{{Method}} 
         &  \multicolumn{1}{c}{CLIP ($\rm N$-$\rm I$) $\uparrow$}
         \\
         
         \midrule    
         w/ Spatially Decoupled Attention
         & \textbf{85.37}
         \\

         w/o Spatially Decoupled Attention
         & 81.28
         \\
         \bottomrule
    \end{tabular}
    \caption{
    Quantitative analysis of Janus Artifacts, measuring a CLIP score between rendered normals and textures across random views.
    }
    \label{table:janus}
\end{table}

\subsection{Analysis of the Occlusion}
In this section, we provide an in-depth analysis of occlusion inpainting to address limitations in our main method and clarify its comparative strengths and weaknesses. While our paper demonstrates the effectiveness of our 3D-aware occlusion inpainting approach across diverse scenarios, there are complex cases that challenge even the most advanced techniques. 
This supplementary section highlights these edge cases and evaluates the relative performance of three methods: our 3D-aware inpainting, automatic view selection~\cite{chen23text2tex}, and UV ControlNet~\cite{zeng24paint3d}. 
By examining these approaches, we aim to shed light on the specific strengths and limitations of our method in comparison to alternatives, ultimately guiding future improvements.
\begin{figure*}[b]
    \centering
    \includegraphics[width=0.85\linewidth]{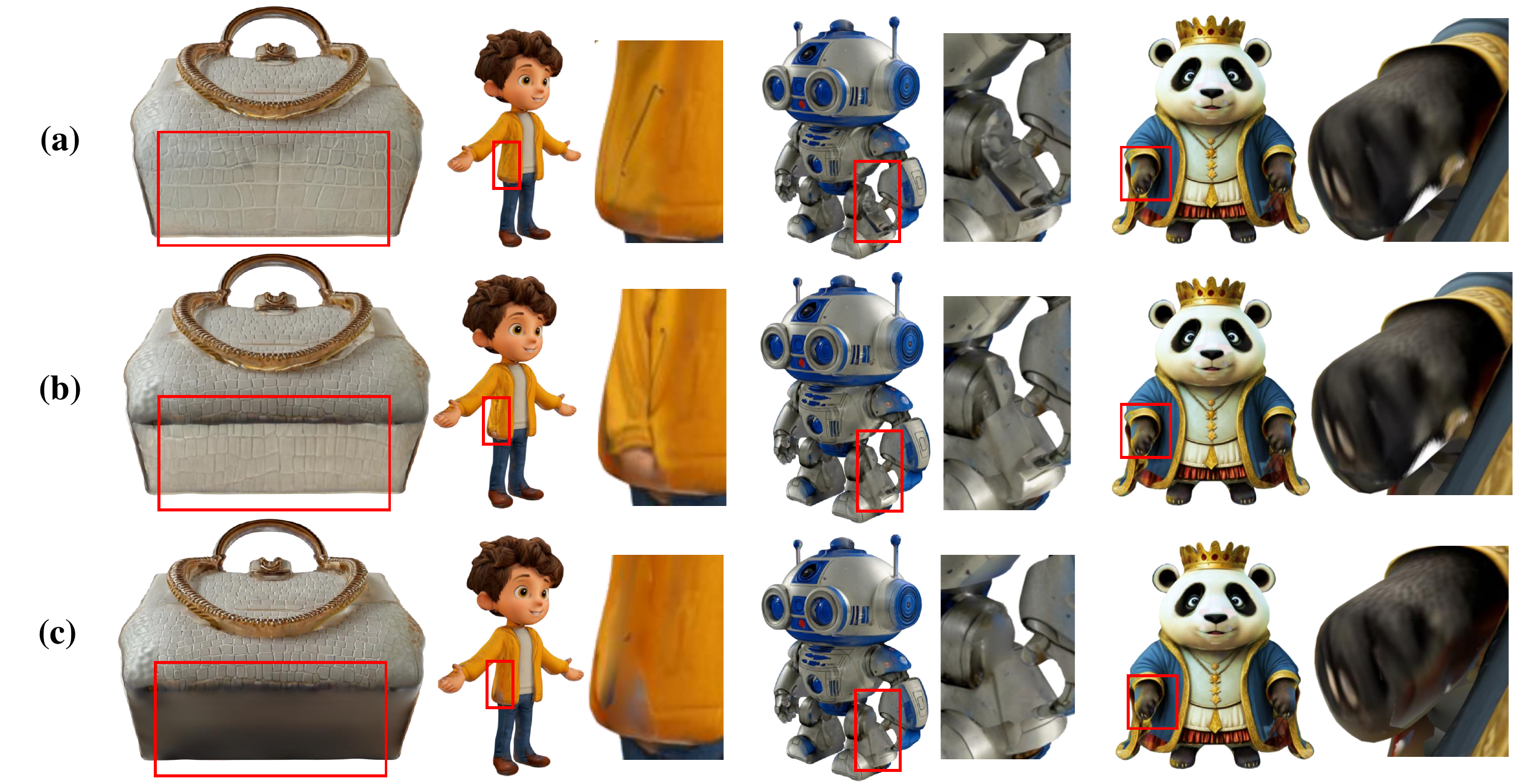}
    \caption{\textbf{Qualitative results for different occlusion inpainting methods}.
    (a) shows results from our 3D-aware occlusion inpainting method, (b) uses automatic view selection, and (c) employs UV ControlNet. 
    }
    \label{fig:occ_edge}
\end{figure*}
To explore occlusion handling, we compare (a) our proposed 3D-aware occlusion inpainting, (b) automatic view selection, and (c) UV ControlNet, with an emphasis on texture fidelity, seam visibility, and processing efficiency.

As seen in Figure~\ref{fig:occ_edge}, both our method and automatic view selection exhibit high fidelity and maintain strong texture coherence, effectively blending textures with the surrounding context in most instances. This results in fewer visible seams, a crucial factor in generating visually coherent 3D assets. 
However, when faced with certain edge cases, such as complex or highly occluded areas (\textit{e.g.,} the panda example), both methods (a) and (b) encounter challenges in filling occluded regions.

On the other hand, as demonstrated in Figure~\ref{fig:occ_edge} (c), UV ControlNet shows its capability to fill all occlusions comprehensively in UV space, albeit at a reduced fidelity level. 
This trade-off reflects the inherent limitations of 3D-aware methods and automatic view selection when attempting to achieve high coverage in extreme cases. 
UV ControlNet achieves full occlusion fill by directly addressing gaps in the UV space, albeit with a reduction in texture detail, which may not be ideal for high-fidelity applications. 
In contrast, our 3D-aware inpainting offers superior texture fidelity and coherence across most scenarios.
This trade-off between coverage and detail emphasizes the potential for future enhancements that balance high-quality inpainting with comprehensive occlusion handling.

\begin{figure*}[h]
    \centering
    \includegraphics[width=0.94\linewidth]{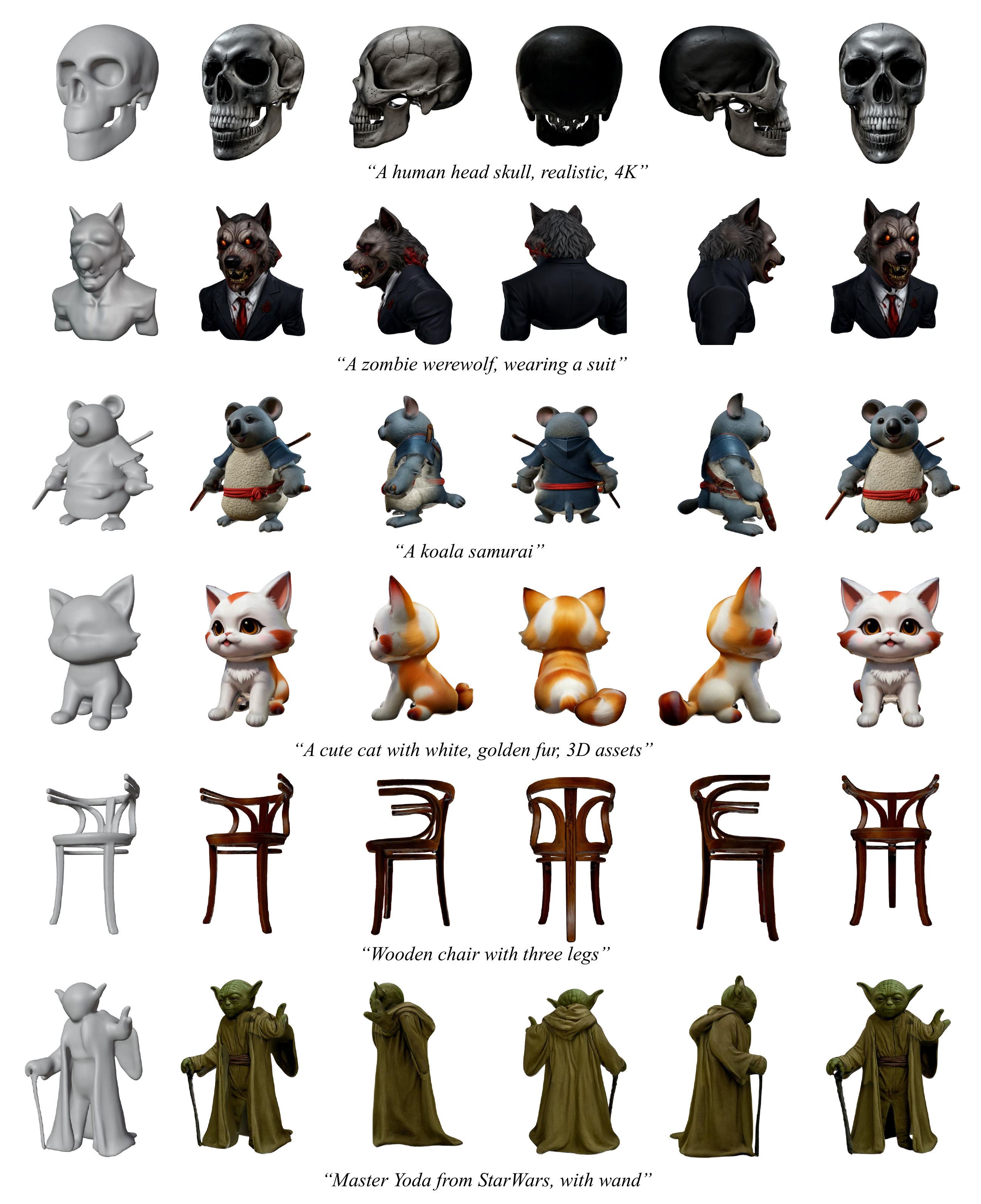}
    \caption{\textbf{Text-to-3D Results of CaPa}.
    CaPa can generate diverse objects from textual, and visual input. 
    The result underscores CaPa's strengths in generating high-resolution textures that align with well-defined geometries.
    }
    \label{fig:text1}
\end{figure*}
\subsection{Text-to-3D Results of CaPa}
Lastly, Figure~\ref{fig:text1} demonstrates CaPa's capabilities in generating high-quality 3D assets from textual prompts. 
Results across different object categories, including characters, common objects, and cultural artifacts, affirm CaPa’s generalization ability to maintain geometric stability and texture consistency in various shapes and surface types.
\section{Additional Details of CaPa}
\label{app:detail}
\subsection{Mesh Extraction and Remeshing Algorithm}
\label{app:mesh}
In 3D asset generation, obtaining a high-quality mesh is essential for downstream applications, as it allows the asset to be integrated effectively into various environments, such as rendering engines and simulation pipelines. 
While recent advancements in Neural Rendering, such as NeRF and Gaussian Splatting, these representations suffer from accurate surface reconstruction~\cite{tang2022nerf2mesh, guedon24sugar}. 
As a result, extracting meshes from Gaussian Splatting or NeRF degrades mesh quality after a typical mesh conversion algorithm like Marching Cube~\cite{marchingcube}.

In response, DMTet~\cite{dmtet} and FlexiCubes~\cite{flexicube} have addressed this problem by optimizing meshes through external learnable parameters for meshing. 
While these approaches significantly improve output mesh quality, the computational cost is intensive and still generates incomplete or suboptimal geometries. 
\subsubsection{Initial Mesh Extraction}
Unlike previous works with NeRF or Gaussian Splatting representation, our 3D diffusion model operates on occupancy fields that inherently align better with mesh extraction methods. 
This compatibility enables us to achieve cleaner, higher-fidelity geometry, avoiding the substantial quality loss common in meshing processes for NeRF and GS-based assets.
We first extract the initial mesh of the generated occupancy field by leveraging the Marching Cube algorithm~\cite{marchingcube}. 
\newline

\subsubsection{Automated Remeshing Post-Processing}
After the mesh conversion, we design an automated post-processing pipeline to obtain a cleaner topology.
This post-processing enhances the geometric quality and accelerates the extraction process.
Our pipeline begins by ensuring manifold properties of the generated mesh through a series of cleaning and remeshing steps. 
The mesh cleaning process enhances quality by removing unreferenced or isolated elements and repairing non-manifold components. 
Floater removal processes unreferenced vertices by merging duplicates and removing zero-area faces as well as small, isolated components. 
For non-manifold repair, faces connected to non-manifold edges are iteratively removed, starting with the smallest faces, to maintain 2-manifold properties.
Our remeshing step offers flexibility by supporting both triangular and quadrilateral mesh options. 
\newline

\noindent{\textbf{Triangular Remeshing.}}
For {triangular remeshing}, following Botsch et al.~\cite{triremesh}, we iteratively refine the base mesh through isotropic remeshing to form equilateral triangles. 
This process involves alternating edge lengths and vertex valence equalization to generate a uniformly remeshed surface. 
Specifically, edge lengths are adjusted to a predefined target constant through edge splitting and collapsing, achieving uniform face sizes across the mesh. 
Then, vertex valence deviation is minimized toward 6 (or 4 at boundaries) by applying edge flips, thereby enhancing structural uniformity. 
Following these steps, an area-based tangential smoothing process is applied to preserve the surface shape. 
Each vertex is adjusted toward a centroid, calculated as a weighted average of neighboring vertices based on area. The gravity-weighted centroid \( \mathbf{g}_i \) for each vertex \( \mathbf{p}_i \) is given by:
\begin{equation}
\mathbf{g}_i := \frac{1}{\sum_{p_j \in N(p_i)} A(p_j)} \sum_{p_j \in N(p_i)} A(p_j) \mathbf{p}_j,
\end{equation}
where \( A(p_j) \) denotes the area associated with vertex \( \mathbf{p}_j \), and \( N(p_i) \) is the set of neighboring vertices of \( \mathbf{p}_i \). This weighted average ensures that vertices with larger areas have a stronger influence, thereby smoothing the mesh more uniformly.

To ensure tangential smoothing on the surface, the update vector is projected back onto the tangent plane of \( \mathbf{p}_i \) as follows:
\begin{equation}
\mathbf{p}_i \leftarrow \mathbf{p}_i + \lambda \left( \mathbf{I} - \mathbf{n}_i \mathbf{n}_i^T \right) (\mathbf{g}_i - \mathbf{p}_i),
\end{equation}
where \( \mathbf{n}_i \) is the normal vector of \( \mathbf{p}_i \), and \( \lambda \) is a damping factor used to avoid oscillations. This projection moves each vertex in the tangent direction toward its centroid, smoothing the mesh while preserving its overall shape.
\newline

\noindent{\textbf{Quadrilateral Remeshing.}} 
Following Jakob et al.~\cite{quadremesh}, quadrilateral remeshing is designed to produce structured meshes composed of quadrilateral faces, offering a uniform and easily manageable layout ideal for various applications requiring high-quality surface control. 
This process is accomplished through two primary stages: 1) Orientation Field Optimization and 2) Position Field Optimization.
\newline

{\textit{1) Orientation Field Optimization:}} 
This initial step establishes directional alignment across the mesh surface by defining an orientation field. This field prescribes specific directional vectors at each vertex, effectively creating a structured “grid” on the surface. By aligning each edge along these pre-defined directions, the mesh gains a cohesive pattern that ensures smooth transitions across faces, an essential property for generating a clean quadrilateral structure.
\newline

{\textit{2) Position Field Optimization:}} 
Once edge orientations are set, the positions of vertices are optimized to achieve consistent spacing between edges, regulated by a global scale factor. This scale factor, provided by the user, enforces uniform edge lengths across the entire mesh, ensuring that each quadrilateral face remains similarly sized. Vertices are locally adjusted on a 2D tangent plane to fit within this structured layout, resulting in a smooth, evenly distributed quadrilateral grid.
\newline

\noindent{By} combining these two stages, the quadrilateral remeshing process yields a quadrilateral mesh that maintains both geometric coherence and structural regularity, making it suitable for applications demanding high fidelity and uniform surface representation. This approach allows for a stable, grid-like arrangement that balances surface alignment with positional consistency across the mesh.
For further details on quadrilateral remeshing, we refer readers to the Instant Field-Alignment~\cite{quadremesh}.
\newline

\begin{figure}[h]
    \centering
    \includegraphics[width=0.8\linewidth]{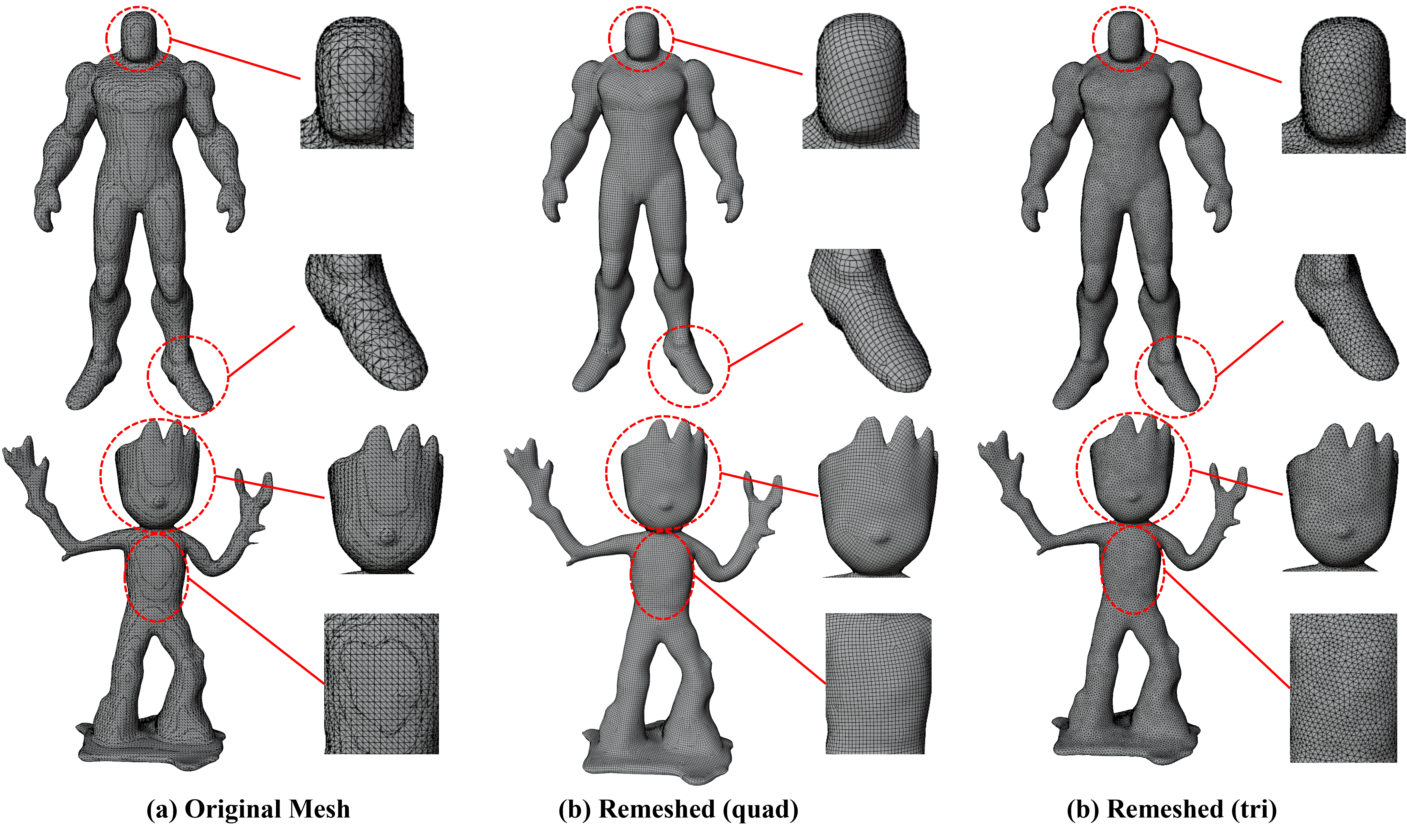}
    \caption{\textbf{Results of Our Remeshing Algorithm}.
    We employ a carefully designed remeshing scheme after geometry generation for better practical usage for broader applications.
    (a) shows the original polygonal mesh, (b) shows remeshed output of quadrilateral faces, and (c) shows remeshed output of triangular faces.
    }
    \label{fig:remesh}
\end{figure}

\noindent{\textbf{Experimental Results.}} 
Figure~\ref{fig:remesh} illustrates the results before and after remeshing. 
In Figure~\ref{fig:remesh} (b), the triangle remeshing outcome yields a highly regular mesh compared to Figure~\ref{fig:remesh} (a), with uniformly distributed Voronoi regions around each vertex. 
Similarly, as shown in Figure~\ref{fig:remesh} (c), each quadrilateral face is uniform, and the overall mesh contains fewer irregular vertices. These results demonstrate how post-processing forms a smooth and stable mesh structure.

In conclusion, our automated post-processing pipeline improves the uniformity and quality of generated meshes while significantly reducing processing time. Our approach yields visually superior, computationally efficient meshes through manifold enforcement, flexible remeshing techniques, and advanced texture unwrapping, optimizing the overall 3D asset generation workflow.

\subsection{Texture Generation}
\label{app:texture_synthesis}
As mentioned in the main manuscript, we first render four orthogonal views of the generated mesh for texture synthesis. 
Then, we generate corresponding images for rendered inputs, by leveraging our novel spatially decoupled attention mechanism.
In this stage, to achieve a state-of-the-art quality of texture generation, we have implemented our texture generative scheme on the SDXL~\cite{podell2024sdxl}. 
We have utilized pre-trained IP Adapter~\cite{ye2023ip-adapter} layer and CLIP~\cite{radford21clip} for the image feature extractor, Depth-ControlNet~\cite{zhang23controlnet} for a geometric guide to align with the geometry.
Notably, to generate texture efficiently, we have adopted the SDXL lightning model~\cite{lin2024sdxllightningprogressiveadversarialdiffusion} baseline checkpoint of our texture generation and occlusion inpainting.
With this diffusion pipeline, we have operated only 8 steps per inference with DPM-solver. 

However, at this stage, directly applying textures from four views often results in visible seams where the views meet.
In response, we leveraged a smoothing mechanism specifically for the side views, similar to the previous works~\cite{chen23text2tex, bensadoun2024meta3dtexturegenfast, kim2024rocotex}. 
When generating the side textures, we mask out the regions already visible in the front and back views, using a confidence map that softly transitions the mask. 
Here, the confidence map is calculated by measuring the cosine similarity between the viewing direction and the normal vector of the mesh faces.
We apply differential denoising during texture synthesis, which only partially influences the masked regions. 
The confidence map controls the strength of the effect. 
This allows the side textures to blend smoothly with the front and back views.
\newline

\noindent{\textbf{Remarks for Occlusion Inpainting.}} 
It is important to note that the occlusion-specific UV map is only used for inpainting purposes. As this map captures a limited set of projections, it may introduce artifacts such as visible seams if used for final texturing. To generate the actual texture map, we utilize Blender's smart UV projection \cite{Blender}, to ensure a consistent and high-quality texture across the entire mesh surface.

\end{document}